\documentclass[runningheads]{llncs}
\usepackage{graphicx}

\usepackage{tikz}
\usepackage{comment}
\usepackage{amsmath,amssymb} 
\usepackage{color}
\usepackage{marvosym}
\usepackage[accsupp]{axessibility}  

\usepackage{array}
\usepackage{multirow}
\usepackage{colortbl}
\usepackage{cite}
\usepackage[ruled]{algorithm2e}
\usepackage{enumitem}
\usepackage{bm}
\usepackage{bbm}
\usepackage{gensymb}
\usepackage{wrapfig}
\usepackage{lipsum}

\newcommand{\tabref}[1]{Tab.~\ref{#1}}

\newcommand\G[1]{\textcolor{green}{#1}}
\newcommand\B[1]{\textcolor{blue}{#1}}

\makeatletter
\newcommand{\thickhline}{%
    \noalign {\ifnum 0=`}\fi \hrule height 1pt
    \futurelet \reserved@a \@xhline
}

\newcommand{\ie}{\textit{i}.\textit{e}.}
\newcommand{\eg}{\textit{e}.\textit{g}.}
\makeatother

\begin{document}
\pagestyle{headings}
\mainmatter
\def\ECCVSubNumber{3655}  

\title{Semi-supervised 3D Object Detection with Proficient Teachers} 

\titlerunning{ProficientTeachers}

%
\author{Junbo Yin \inst{1*} \and
Jin Fang \inst{2,3,4*} \and
Dingfu Zhou \inst{2,3} \and
Liangjun Zhang \inst{2,3} \and \\
Cheng-Zhong Xu \inst{4} \and
{\Letter}Jianbing Shen \inst{4}  \and
{\Letter}Wenguan Wang \inst{5} 
}
\authorrunning{J. Yin et al.}
%
\institute{$^1$School of Computer Science, Beijing Institute of Technology \quad
$^2$Baidu Research \\
\mbox{$^3$National Engineering Laboratory of Deep Learning Technology and Application, China} \\
\mbox{$^4$SKL-IOTSC, CIS, University of Macau \,
$^5$ReLER, AAII, University of Technology Sydney}
\\
\email{\{yinjunbocn,wenguanwang.ai\}@gmail.com}
\quad
\url{https://github.com/yinjunbo/ProficientTeachers}
} 

\maketitle

\begin{abstract}
Dominated point cloud-based 3D object detectors in autonomous driving scenarios rely heavily on the huge amount of accurately labeled samples, however, 3D annotation in the point cloud is extremely tedious, expensive and time-consuming. To reduce the dependence on large supervision, semi-supervised learning (SSL) based approaches have been proposed. The Pseudo-Labeling methodology is commonly used for SSL frameworks, however, the low-quality predictions from the teacher model have seriously limited its performance. In this work, we propose a new Pseudo-Labeling framework for semi-supervised 3D object detection, by enhancing the teacher model to a proficient one with several necessary designs. First, to improve the recall of pseudo labels, a Spatial-temporal Ensemble (STE) module is proposed to generate sufficient seed boxes. Second, to improve the precision of recalled boxes, a Clustering-based Box Voting (CBV) module is designed to get aggregated votes from the clustered seed boxes. This also eliminates the necessity of sophisticated thresholds to select pseudo labels. Furthermore, to reduce the negative influence of wrongly pseudo-labeled samples during the training, a soft supervision signal is proposed by considering Box-wise Contrastive Learning (BCL). The effectiveness of our model is verified on both ONCE and Waymo datasets. For example, on ONCE, our approach significantly improves the baseline by 9.51 mAP. Moreover, with half annotations, our model outperforms the oracle model with full annotations on Waymo.

\end{abstract}

\let\thefootnote\relax\footnotetext{{\Letter}: Corresponding author. \\
{*}: Equal contribution. Work done when J. Yin was an intern at Baidu Research.}

\section{Introduction} \label{sec:intro}

With the rapid development of range sensors (\eg, LiDAR) and their wide application in the field of robotics and autonomous driving, point cloud-based scene understanding such as 3D object detection has received great attention recently. With the great capabilities of deep neural networks (DNNs) and a huge number of annotated samples, impressive performances have been achieved on different public benchmarks \cite{geiger2012we, WAYMO_2020Sun,caesar2019nuscenes, Apolloscape_2019_huang}. VoxelNet \cite{VoxelNet_2018_Zhou}, PointRCNN \cite{PointRCNN_2019SHI}, PointPillars \cite{PointPillars_2018LANG}, PV-RCNN \cite{PVRCNN_2020_Shi} and CenterPoint \cite{CENTERPOINT_2021} are several representative 3D object detection frameworks. Nevertheless, the results highly rely on huge annotations, while the annotation in 3D data is extremely expensive and time-consuming, \eg, a skilled worker may take hundreds of hours to 
annotate just one hour of driving data~\cite{mao2021one,meng2021towards,LIDAR_SIM_2020_fang,meng2020weakly}.

\begin{figure}[t]
\includegraphics[width=0.95\textwidth]{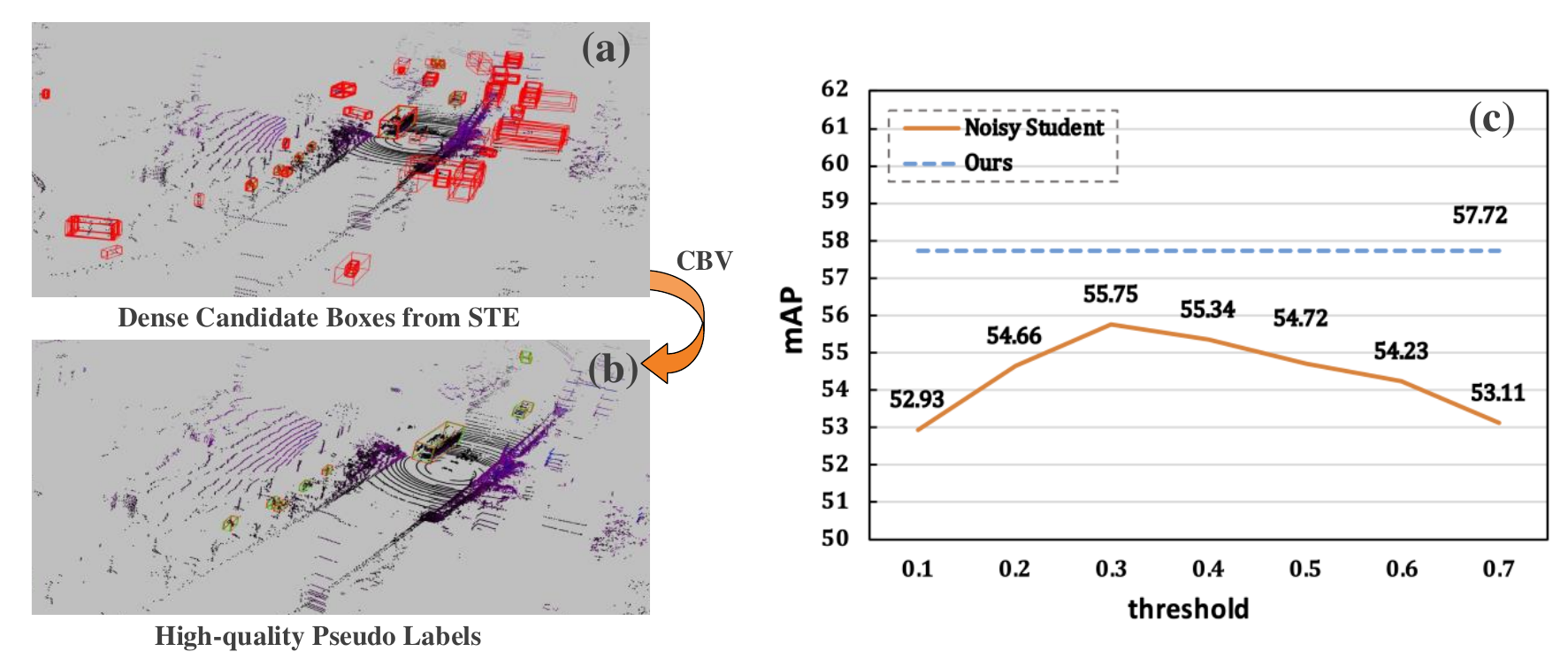}
\caption{\textbf{The main idea of our \textit{ProficientTeachers}.} Given an unlabeled point cloud, the spatial-temporal ensemble (STE) module first produces \textbf{(a)} seed boxes by combing predictions from multiple augmented views. Then, the clustering-based box voting (CBV) module adaptively aggregates these boxes to get \textbf{(b)} the final high-quality pseudo labels. In this way, we not only achieve better detection results, but also remove the necessity of 
sophisticated thresholds for selecting pseudo labels, as seen in \textbf{(c)}.}

\label{fig:threshold}
\end{figure}

Semi-supervised learning (SSL) techniques, which train a model with a small number of labeled samples together with an abundance of unlabeled data, are a promising alternative to the fully-supervised learning frameworks. Compared to labeled data, unlabeled data is obtained very conveniently and cheaply. However, due to the inherent difficulty of point cloud (\eg, orderless, textureless, and sparsity), only a few SSL-based 3D object detection frameworks have been proposed.
Up to now, SESS~\cite{SESS_2020} and 3DIoUMatch~\cite{3DIoUMatch_2021} are two pioneers of this domain. To handle the unlabeled data, SESS leverages asymmetric data augmentation and enforces consistency regularization between the predictions of teacher and student models. Although noticeable improvements have been achieved upon a vanilla VoteNet \cite{DeepVoteNet_2019} on indoor datasets, other researchers \cite{3DIoUMatch_2021,rizve2021defense} found that the consistency regularization is suboptimal if it is uniformly enforced on all the student and teacher predictions because the quality of these predictions may be quite different. To well handle this limitation, 3DIoUMatch seeks a pseudo-labeling approach and applies a confidence-based filtering strategy for pseudo-label selection, where the confidence is defined as a combination of classification score and IoU estimation. Though achieving better performance, it takes tremendous effort to select a suitable confidence score. Moreover, the pseudo labels produced by its plain teacher model limit the final detection performance.

To address these challenges, we propose a new pseudo-labeling SSL framework, \textit{ProficientTeachers}, that not only provides high-quality pseudo labels via an enhanced teacher model, but also reduces the necessity of deliberately selected thresholds. To be specific, false negative (FN) and false positive (FP) in the pseudo labels are two main challenges. The LiDAR point cloud is sparse and noisy, and some street objects are of small sizes (e.g., less than $2m$ for pedestrian) and unevenly distributed across a considerably wide range (e.g., $150m\times150m$ in Waymo~\cite{WAYMO_2020Sun}). Thus it is prone to cause FN detections from only a single point cloud view. To handle this, a spatial-temporal ensemble (STE) module is proposed to generate sufficient seed boxes from spatially and temporally augmented views. Aggregating predictions from different views reduces the prediction bias of the teacher model, thus essentially boosting the \textit{recall}. The redundant seed boxes from STE inevitably involve FP detections. To further resolve the FP problem, we propose a clustering-based box voting (CBV) module. Our CBV module groups the seed boxes into different clusters and generates votes (\ie, a refined bounding box) for each box in a cluster. These votes are then aggregated to produce a more accurate box for each cluster. In this way, more clean and precise pseudo labels can be obtained by a simple NMS, removing the need of selecting thresholds. Our CBV module significantly improves the \textit{precision} of pseudo labels produced by the STE module, without losing the \textit{recall}. By equipping the STE and CBV modules, the vanilla teacher model has become to be {proficient teachers} model. Furthermore, we find that the original pseudo-labeling method enforces a hard training target, where the inaccurate pseudo labels will undermine the performance of the student. To alleviate this problem, a soft training target is proposed by box-wise contrastive learning (BCL), which aims to learn the cross-view feature consistency based on the informative boxes.

To summarize, we propose a new 3D SSL framework, \textit{ProficientTeachers}, for LiDAR-based 3D object detection, which is achieved by promoting the plain teacher model to {proficient teachers} inspired by ensemble learning. Our framework not only performs better results, but also removes the necessity of confidence-based thresholds for filtering pseudo labels. In our model, spatial-temporal ensemble (STE) and clustering-based box voting (CBV) modules are developed to improve the recall and precision of pseudo labels. Furthermore, a box-wise contrastive learning (BCL) strategy has been advocated to explore representation learning based on expressive 3D boxes. Comprehensive evaluations have been conducted on ONCE and Waymo. Our \textit{ProficientTeachers} can improve the baseline detector by 9.51 mAP on ONCE, and save half annotations on Waymo.
\section{Related Work}\label{sec:related_work}

 \noindent\textbf{3D Object Detection.} 3D object detection from LiDAR point cloud has been studied for decades with various approaches being proposed. The mainstream methodologies can be divided into two categories: voxel-based~\cite{VoxelNet_2018_Zhou,yin2021graph,yin2020lidar,CENTERPOINT_2021,Multi_view_CNNSEG_2017_CHEN} and point-based~\cite{PointRCNN_2019SHI,Joint_3D_Instance_2020_Zhou,shi2020point,yang20203dssd}. Voxel-based expression is a popular way of processing the point cloud in deep learning, and has been widely applied for 3D object detection with the development of sparse convolution \cite{Second_2018Yan}. Voxelnet \cite{VoxelNet_2018_Zhou} splits the LiDAR data into voxels and sends the points in each voxel into a voxel feature encoding layer to get voxel-wise features. To accelerate the speed, PointPillars~\cite{PointPillars_2018LANG} uses a pillar representation to replace the voxel representation. Different from the above approaches, the point-based approaches take the point cloud directly into the DNNs. PointRCNN~\cite{PointRCNN_2019SHI} is representative of this, which employs the Pointnet++ \cite{PointNet++_2017} as the backbone for semantic segmentation first, and then regions of interest (RoIs) are generated based on foreground points. Besides this, PV-RCNN~\cite{PVRCNN_2020_Shi} proposes to combine both point cloud and multi-scale voxel representation and achieve high performance.  
 

\noindent\textbf{Semi-supervised Learning (SSL).} SSL has been studied for a long time and many approaches have been proposed \cite{Introduction_SSL_2009_ZHU, SSL_SURVEY_2020}. 
The recently popular SSL approaches such as {Temporal Ensembling} \cite{Temporal_Ensembling_2017_Samuli}, {Mean Teacher} \cite{Mean_Teacher_2017_Tarvainen} and  Noisy Student \cite{Noisy_student_2020} have achieved impressive performance on the 2D tasks. The temporal model, which explores consistency in the prediction level, tries to minimize the difference between the predictions from the current step and the EMA (an exponential moving average) predictions over multiple previous training epochs. The EMA predictions can largely improve the quality of the predictions.
Mean Teacher \cite{Mean_Teacher_2017_Tarvainen} further improves the temporal model by replacing network prediction average with network parameter average. It contains two network branches, \ie, teacher and student, with the same architecture. The parameters of the teacher are the EMA of the student, while the parameters of the student are updated by stochastic gradient descent. The student network is trained to yield consistent predictions with the teacher network. Noisy Student \cite{Noisy_student_2020} deliberately injects noise to the student model from both input level and model level to strengthen the student. Then it makes the student a new teacher and performs iterative self-training to further improve performance. 

\noindent\textbf{SSL on 3D Object Detection.} Most of the SSL approaches are proposed for classification tasks \cite{MixMatch_2019mixmatch, FixMatch_2020fixmatch} and a few SSL approaches have been proposed to leverage the object detection task \cite{sohn2020simple, CSD_2019consistency}, especially 3D object detection. SESS \cite{SESS_2020} and 3DIoUMatch \cite{3DIoUMatch_2021} are two typical recently proposed approaches for 3D object detection from point cloud data. The SESS, which is built based on the Mean Teacher paradigm by updating the parameters of the teacher network with an EMA technique, employs asymmetric data augmentation and enforces three kinds of consistency losses between the teacher and student predictions. Different from SESS, 3DIoUMatch proposes a series of handcrafted strategies to achieve better pseudo labels such as joint class, objectness, localization confidences-based pseudo-label filtering, and IoU-guided lower-half suppression for deduplication, etc. The proposed framework works well on PV-RCNN \cite{PVRCNN_2020_Shi}. By contrast, our method benefits from an enhanced teacher model, without sophisticated thresholds selection, and can be flexibly applied on popular LiDAR-based 3D detectors, \eg, SECOND~\cite{Second_2018Yan}, CenterPoint~\cite{CENTERPOINT_2021} and PV-RCNN~\cite{PVRCNN_2020_Shi}.




\section{Proposed Approach}\label{sec:proposed}


\begin{figure}[t]
\begin{center}
\includegraphics[width=0.95\textwidth]{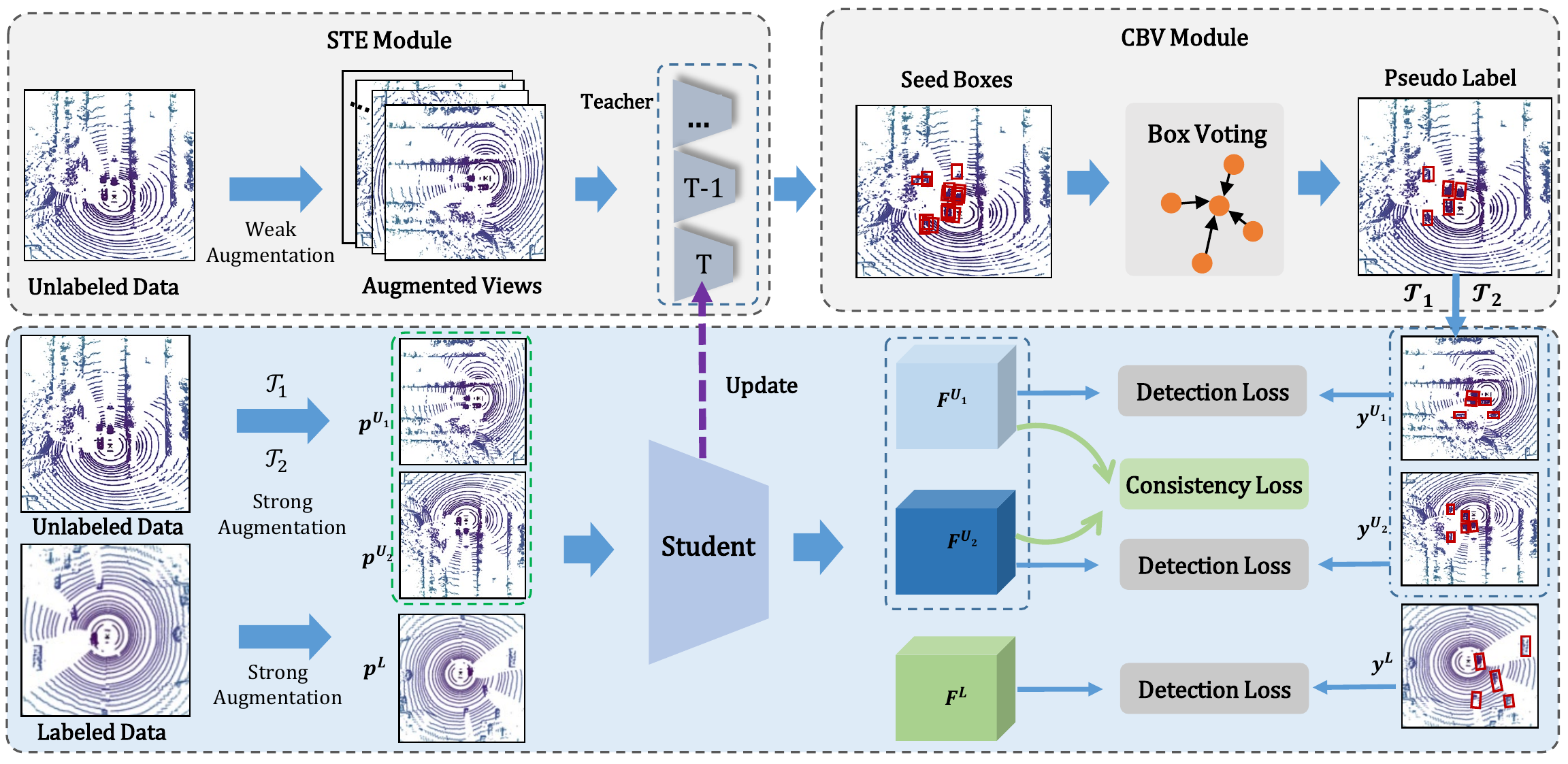}
\end{center}
\caption{\textbf{The framework of our \textit{ProficientTeachers} model.} It consists of a spatial-temporal ensemble (STE) module and a clustering-based box voting (CBV) module. STE produces sufficient seed boxes and CBV adaptively fuses them to obtain better pseudo labels. Besides, a consistency loss based on box-wise contrastive learning (BCL) is applied on the student model to explicitly learn from informative box features.}
\label{fig:framework}
\end{figure}

\subsection{SSL Framework} \label{subsec:framework}
Semi-supervised 3D object detection is very valuable and practical in real self-driving scenarios. Here, we explore the {pseudo-labeling} framework, which is a popular methodology in semi-supervised learning. Typical {pseudo-labeling} methods~\cite{Noisy_student_2020} exploit a teacher model to produce pseudo labels which are used to supervise a student model. We argue that the low-quality predictions produced by the vanilla teacher model limit the performance. In this work, we propose to promote the vanilla teacher model to a proficient one, as seen in Fig.~\ref{fig:framework}.

Our \textit{ProficientTeachers} model, as described in Sec.~\ref{subsec:teacher}, contains a spatial-temporal ensemble (STE) module and a clustering-based box voting (CBV) module, which are designed to handle false negative (FN) and false positives (FP) in pseudo labels, respectively. In particular, STE produces multi-group boxes based on augmented and assembled views to recall the missed objects. To further remove the redundant FP boxes, the original {pseudo-labeling} approaches apply a fixed threshold to filter out boxes with lower confidence scores, which is unstable and inefficient. In contrast, our CBV adaptively aggregates the seed boxes to reduce FP and also improves the precision by voting the boxes within a cluster. Furthermore, we propose a soft supervision signal in Sec.~\ref{subsec:student}, \ie, \textit{Contrastive Student} model, by involving box-wise contrastive learning (BCL). Next, we elaborate on each module in subsequent sections.

\subsection{Proficient Teachers Model}
\label{subsec:teacher}


Assuming that we have total $N$ training samples, including $N_l$ labeled samples $\mathcal{P}^{L} = \{\mathbf{p}_i^{L}, \mathbf{y}_{i}^{L}\}_{i=1}^{N_l}$ and $N_u$ unlabeled samples $\mathcal{P}^{U} = \{\mathbf{p}_i^{U}\}_{i=1}^{N_u}$, where $\mathbf{p}_{i} \in \mathcal{R}^{n \times \{3+r\}}$ represents a point cloud sample $\mathbf{p}_i$ that has $n$ points with 3-dimensional coordinates and other $r$-dimensional attributes such as intensity, timestamps, etc. The core idea of our \textit{ProficientTeachers} model is to produce high-quality pseudo labels $\{\mathbf{y}_{i}^{U}\}_{i=1}^{N_u}$ for the unlabeled point clouds $\{\mathbf{p}_i^{U}\}_{i=1}^{N_u}$.  

\noindent\textbf{Spatial-Temporal Ensemble Module.} As described above, predictions produced by vanilla teacher inevitably encounter the FN problem, which will cause the student model to treat a foreground object as a negative example, thus degenerating the detection performance. Motivated from this observation, we attempt to improve the recall of the teacher model with the STE module, which generates sufficient candidate pseudo boxes with a spatial ensemble module and a temporal ensemble module.

The spatial ensemble in the STE aims to produce multi-group detections based on differently augmented point cloud views, which is inspired by test time augmentation~\cite{shanmugam2021better}. Detections on a certain view may exist bias and miss objects. By combining detections from multiple views after reverse transformation, less FN will happen, thus improving the recall. Formally, given an unlabeled data $\mathbf{p}_{i}^{U}\in\mathcal{P}^{U}$, we first apply \textit{fixed} data augmentation to spatially transform the input point cloud into different views. This can be deemed as a form of weak augmentation compared with the \textit{random} train-time augmentation policies, which is: 
\begin{equation}
\{\mathbf{p}_{i}^{U_1}, \mathbf{p}_{i}^{U_2}, \ldots, \mathbf{p}_{i}^{U_K}\}=\mathcal{T}(\mathbf{p}_{i}^{U}),
\end{equation}
where $\mathcal{T}(\cdot)$ is the augmentation function and $K$ is the number of augmented views. More details about our data augmentation can be found in Sec.~\ref{subsec:detail}.

Then, we use the teacher network $f_\text{TEA}(\cdot)$ to give predictions for each view:  
\small
\begin{equation}
\{\mathbf{y}_{i}^{U_1}, \mathbf{y}_{i}^{U_2}, \ldots, \mathbf{y}_{i}^{U_K}\}=f_\text{TEA}(\{\mathbf{p}_{i}^{U_1}, \mathbf{p}_{i}^{U_2}, \ldots, \mathbf{p}_{i}^{U_K}\}),
\end{equation}
\normalsize
With the spatial ensemble module, objects not detected in one view may be detected in another view. All detection results are then projected to the original coordinate with reverse transformations.

Moreover, we can also incorporate a temporal ensemble module. In particular, previous pseudo label methods update weight of teacher model by student model from a certain epoch. We argue that models from different epochs exist bias, and temporal aggregation of models from different epochs can alleviate this problem. This is presented as: 
\small
\begin{equation}
\{\mathbf{y}_{i}^{U_1^{T}}, \mathbf{y}_{i}^{U_2^{T}}, \ldots, \mathbf{y}_{i}^{U_K^{T}}\}=f_\text{TEA}^{T}(\{\mathbf{p}_{i}^{U_1}, \mathbf{p}_{i}^{U_2}, \ldots, \mathbf{p}_{i}^{U_K}\}),
\end{equation}
\normalsize  
where $f_\text{TEA}^{T}(\cdot)$ denotes the teacher model updated by student model from epoch $T$. We consider the models from recent epochs for temporal ensemble. Afterward, we collect all the predictions produced by both the spatial and temporal ensemble modules. 
As seen in Table~\ref{tab:recall_pre}, our STE improves the recall by 5.6\%, compared with the original teacher model. Here, simple post-processing like Non-maximum Suppression (NMS) can be directly applied to aggregate these predictions. 




\begin{figure}[t]
\begin{center}
\includegraphics[width=0.98\textwidth]{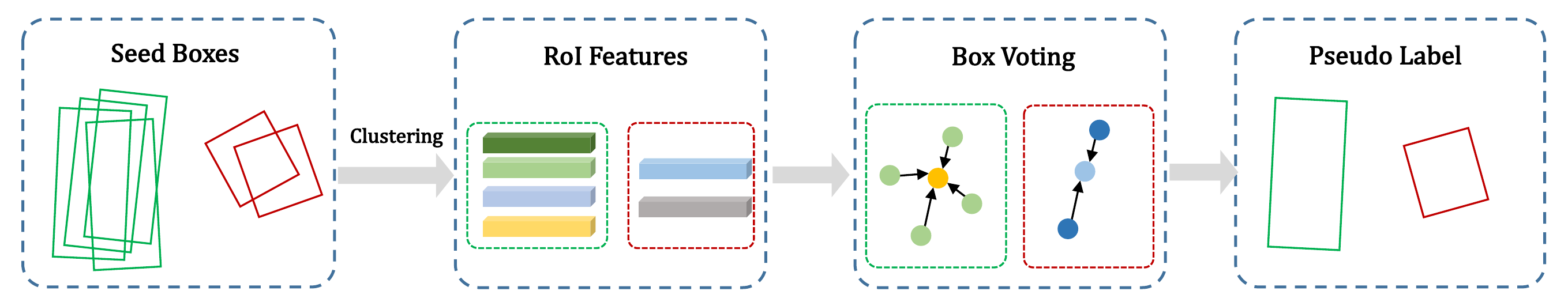}
\end{center}
\caption{\textbf{Illustration of clustering-based box voting (CBV) module}, which contains four steps: seed box clustering, RoI feature extraction, box voting and votes aggregation. By adaptively fusing the high-recall candidate boxes output by the STE, CBV significantly improves the precision of pseudo labels.}
\label{fig:framework_box}
\end{figure}

\noindent\textbf{Clustering-based Box Voting Module.} 
Though the proposed STE module has ensured a higher recall by combing sufficient pseudo boxes, some inaccurate candidate boxes will also lead to a lower precision (\ie, from 47.2\% to 27.4\%, as seen in Table.~\ref{tab:recall_pre}), which undermines the quality of pseudo labels. Thus, the main purpose of the CBV module is to address these redundant candidate boxes.
\begin{wraptable}{r}{6cm}
	\centering
	\caption{Recall and precision of ``Vehicle'' class based on SECOND detector trained with different SSL settings. ``c'' represents the threshold of the confidence score.}
\label{tab:recall_pre}
\resizebox{0.48\textwidth}{!}{%
\begin{tabular}{c|lll} 
\hline
Methods & Recall & Precision  \\ \hline
SECOND \cite{Second_2018Yan} (Baseline) & 78.3\% & 47.2\% \\  \hline
SECOND \cite{Second_2018Yan} + STE & 83.9\% & 27.4\%  \\
SECOND \cite{Second_2018Yan} + STE (c = 0.3) & 80.4\% & 74.6\% \\
SECOND \cite{Second_2018Yan} + STE + CBV & 83.6\% & 76.5\% \\ \hline
\end{tabular}%
}
\end{wraptable}

An intuition way is to define a score to formulate the quality of the pseudo boxes, and filter the low-quality boxes under a certain threshold. For instance,  
FixMatch~\cite{FixMatch_2020fixmatch} directly uses the classification score to filter the pseudo boxes. Later, 3DIoUMatch~\cite{3DIoUMatch_2021} leverages a predicted IoU score for filtering objects. However, there are potential limitations in these threshold-based filtering methods. First of all, it requires manual experience and elaborate experiments 
to choose a suitable threshold for a class, and different datasets or detectors might not share the same thresholds. Secondly, the predicted box confidence score either formulates class probability, \eg, classification score, or the localization accuracy, \eg, IoU, individually, which requires a sophisticated combination from different aspects. Thirdly, they separately consider each individual box, ignoring the relations of boxes in a cluster covering the same ground truth. To this end, we propose a {Clustering-based Box Voting} (CBV) module to obtain more clean and accurate pseudo labels via a learnable voting process, as well as eliminate the need of sophisticated threshold selection.

The detailed pipeline of the CBV module can be found in Fig.~\ref{fig:framework_box}. Specifically, given the pseudo labels produced by the STE module, 
we first cluster these boxes based on the IoU criterion. For example, the boxes are first arranged by the scores. Then, for the box with the highest score, we select other boxes that have a larger IoU with it as the same cluster. This process is iteratively performed for all the boxes to produce all the clusters. Assuming the $k$-th cluster has $M^{'}$ boxes, \ie, $\{\mathbf{b}_{m}^{k}\}_{m=1}^{M^{'}}$, 
we aim to aggregate these boxes and get a refined box $\mathbf{b}^{k}$ via box voting and votes aggregation.
For box voting, we first obtain features of each box
via a pre-trained RoI network~\cite{Deng2021VoxelRT}, \ie, $\tilde{\mathbf{b}}_{m}^{k}\!=\!f_{\text{RoI}}(\mathbf{b}_{m}^{k})$. Then, we let each box in a cluster predict a vote, based on the context-aware features of the box. The voting network is realized by a shared-weight two-layer MLP, and is trained to regress the offset to the ground-truth box. This can be denoted as $\mathbf{v}_{m}^{k}\!=\!f_{\text{Vote}}(\tilde{\mathbf{b}}_{m}^{k})$. In this way, we have the vote results $\{\mathbf{v}_{m}^{k}\}_{m=1}^{M^{'}}$, where each vote $\mathbf{v}_{m}^{k}$ presents a refined bounding box. Furthermore, we also predict an objectness $\mathbf{s}_{m}^{k}$ to describe the quality of a box. This is achieved by training another two-layer MLP head for foreground and background classification. Accordingly, we can aggregate these votes to get a more accurate box $\mathbf{b}^{k}$ for the $k$-th cluster, by weighting the votes $\mathbf{v}_{m}^{k}$ with the objectness $\mathbf{s}_{m}^{k}$, which is denoted as: 
\begin{equation}
\mathbf{b}^{k} = \frac{\sum\nolimits_{m=1}^{M^{'}}{\mathbf{s}_{m}^{k}\cdot \mathbf{v}_{m}^{k}}}{\sum\nolimits_{m=1}^{M^{'}}{\mathbf{s}_{m}^{k}}},
\end{equation}
In this way, more clean pseudo labels are produced. The objectness $\mathbf{s}^{k}$ is also obtained by considering the average scores and the detection number in this cluster. Then, NMS can be directly applied to get the final pseudo labels, \ie, $\mathbf{y}^{U}=\{\mathbf{b}^{k}\}_{k=1}^{M}$, where $M$ is the total pseudo box number.

In Table.~\ref{tab:recall_pre}, we find that the CBV module significantly improves the precision from 27.4\% to 76.5\%, and even improves the recall by 5.3\%, due to the votes aggregation in clusters. We also implement a heuristic method by selecting better confidence threshold, \ie, $c=0.3$, to filter boxes as in~\cite{FixMatch_2020fixmatch}. As seen in Table.~\ref{tab:recall_pre}, such a method is inferior to ours according to both recall and precision metrics.


\subsection{Contrastive Student Model} \label{subsec:student}
In this section, we introduce the learning of the student model. Concretely, our student model takes as input both the labeled data $\{\mathbf{p}^{L}, \mathbf{y}^{L}\}\in\mathcal{P}^{L}$ and the unlabeled data $\mathbf{p}^{U}\in\mathcal{P}^{U}$. For training the labeled data $\mathbf{p}^{L}$, we directly obtain the predictions of the student model, and optimize with the detection loss:  
\begin{equation}
\mathcal{L}^{L}_\text{det}=\mathcal{L}^{L}_\text{cls}+\mathcal{L}^{L}_\text{smooth-$\ell_1$},
\label{eq:totalloss}
\end{equation}
For training the unlabeled data $\mathbf{p}^{U}$, we seek a soft learning target besides the pseudo label-based detection loss. To be specific, though our \textit{ProficientTeachers} model has significantly improved the quality of pseudo labels, it inevitably contains inaccurate predictions. Learning towards these noisy pseudo targets will affect the student performance. To alleviate this problem, we propose to further mine the information in unlabeled data $\mathbf{p}^{U}$, which is achieved by box-wise contrastive learning (BCL). Though contrastive learning~\cite{chen2020simple,xie2020pointcontrast,wang2021exploring,yin2022proposal} has been explored in 3D point cloud, our BCL is different from those by directly contrasting expressive box-level features based on the pseudo predictions.

In particular, our BCL module enforces the feature consistency of the same box instance from different augmented views. Formally, given an unlabeled point cloud $\mathbf{p}^{U}$, we first apply two random augmentations to generate different views $\mathbf{p}^{U_1}$ and $\mathbf{p}^{U_2}$. Then, we get the pseudo labels of the two views:  
\begin{equation}
{\mathbf{y}}^{U_1}=f_\text{STU}(\mathbf{p}^{U_1}), {\mathbf{y}}^{U_2}=f_\text{STU}(\mathbf{p}^{U_1}),
\end{equation} 
where ${{\mathbf{y}}^{U_1}} \in \mathbb{R}^{M_1\times7}$, ${{\mathbf{y}}^{U_2}} \in \mathbb{R}^{M_2\times7}$ and $M_1, M_2$ are the number of the predicted bounding boxes in each view with 7-d attributes. Next, we transform ${{\mathbf{y}}^{U_1}}$ and ${{\mathbf{y}}^{U_2}}$ to the same view, and build positive and negative sample pairs by a greedy matching, \ie, box pairs with the smallest distance are treated as positives and other boxes are viewed as negatives. This results in $M$ matched positive box pairs. Afterwards, we extract box features by a point-wise interpolation method, which is inspired by CenterPoint~\cite{CENTERPOINT_2021}. More precisely, we use the center points features from the six faces of a 3D bounding box to present a box, by applying bilinear interpolation on the bird-eye-view (BEV) feature maps $\mathbf{F}^{U}$. Due to operating on BEV, the center points of the six faces of a 3D box are equal to the center points of the four sides of a 2D box plus one box center. 
This essentially simplifies the box feature extraction process. We formulate this as:
\begin{equation}
\mathbf{h}^{U_1}=I(\mathbf{y}^{U_1}, \mathbf{F}^{U_1}), 
\mathbf{h}^{U_2}=I(\mathbf{y}^{U_2}, \mathbf{{F}}^{U_2}),
\end{equation} 
where $I(\cdot)$ is the bilinear interpolation function, and ${\mathbf{h}}^{U_1}$ and ${\mathbf{h}}^{U_1}$ are resultant box features from different views. These features are then collected and arranged as $\{\mathbf{h}_k\}_{k=1}^{2M}$, where the even indexes denote the boxes from ${\mathbf{h}}^{U_1}$ and the odd ones present boxes from ${\mathbf{h}}^{U_2}$. Later, a projection head $\phi(\cdot)$ that contains two $1\times1$ convolutional layers is used to map the box features to an embedding space, such that $\mathbf{z}_k=\phi(\mathbf{h}_k)$. Then, InfoNCE loss~\cite{oord2018representation} is exploited to build the soft target:  
\begin{equation}
\ell(p,q) = -\log{\frac{\exp({{\mathbf{z}}_p\cdot {\mathbf{z}}_q/\tau})}{\sum\nolimits_{k=1}^{2M}{\mathbbm{1}_{[k\neq{q}]}\exp({{\mathbf{z}}_p\cdot {\mathbf{z}}_k/\tau})}}},
\label{eq:pairloss}  
\end{equation}
\begin{equation}
\mathcal{L}_\text{con}^{U}=\frac{1}{2M}\sum\nolimits_{k=1}^{M}[\ell(2k-1,2k)+\ell(2k,2k-1)],
\label{eq:totalloss}  
\end{equation}
where $p,q\in[1, \ldots, 2M]$ and $p\neq{q}$. $\tau$ is a temperature hyper-parameter that is set to 0.1. Furthermore, we leverage the high-quality pseudo labels produced by the \textit{ProficientTeachers} to construct the detection loss, which is denoted as: 
\begin{equation}
\mathcal{L}^{U}_\text{det}=\frac{1}{2}(\mathcal{L}^{U_1}_\text{cls}+\mathcal{L}^{U_1}_\text{smooth-$\ell_1$}+\mathcal{L}^{U_2}_\text{cls}+\mathcal{L}^{U_2}_\text{smooth-$\ell_1$}).
\label{eq:totalloss}
\end{equation} 
Finally, the overall loss of the student model is defined as: 
\begin{equation}
\mathcal{L}=\mathcal{L}^{L}_\text{det}+\mathcal{L}^{U}_\text{det}+\alpha\mathcal{L}_\text{con}^{U},
\label{eq:totalloss}
\end{equation} 
where $\alpha$ is a coefficient and we set it to 0.05 empirically.

\section{Experimental Results} \label{sec:experiments}


In this work, we first introduce the datasets and the implementation details of our model in Sec.~\ref{subsec:dataset} and Sec.~\ref{subsec:detail}, respectively. Then, we report the main evaluation results by comparing with other SSL approaches in Sec.~\ref{subsec:main_res}. Finally, the ablation studies are presented in Sec.~\ref{subsec:ablation}.

\subsection{Datasets}\label{subsec:dataset}

\noindent\textbf{ONCE Dataset.} ONCE \cite{mao2021one} is a large-scale autonomous driving dataset with 1 million LiDAR point cloud samples. Only 15,000 samples are with annotations which have been divided into training, validation, and testing split with 5K, 3K, and 8K samples, respectively. 
In this dataset, five kinds of foreground objects have been annotated i.e., ``Car'', ``Bus'', ``Truck'', ``Pedestrian'' and ``Cyclist'', while ``Car'', ``Bus'' and ``Truck'' are merged into one class ``Vehicle'' during evaluation. In particular, a specific setting is designed for SSL approaches evaluation, \ie, 5K labeled samples and all the unlabeled samples have been divided into 3 subsets: \textit{Small}, \textit{Medium} and \textit{Large} to explore the effects of different data amounts for SSL-based 3D detection. The small unlabeled set \textit{Small} contains 70 sequences (100k samples), the medium set \textit{Medium} contains 321 sequences (500k samples) and the large set \textit{Large} contains 560 sequences (about 1M samples) in total. Similar to other 3D object detection benchmarks, mean AP (average precision) \cite{everingham2010pascal} over all the classes is employed for evaluation, based on the 3D IoU thresholds 0.7, 0.3 and 0.5 for ``Vehicle'', ``Pedestrian'' and ``Cyclist'', respectively. 
In addition, three different perception ranges, `0-30m'', ``30-50m'', and ``50m-inf'', are specified to well evaluate the performance of 3D detectors.

\noindent\textbf{Waymo Open Dataset.} Waymo~\cite{WAYMO_2020Sun} provides a large-scale LiDAR point cloud dataset that contains 798 sequences (158,361 frames) for training and 202 sequences (40,077 frames) for validation. Since it does not provide additional unlabeled raw data for semi-supervised training. We thus manually tailor a semi-supervised learning dataset following the setting in ONCE~\cite{mao2021one}. Specifically, we divide the 798 Waymo training sequences equally into two splits, \ie, labeled split $\mathcal{P}^{L}$ and unlabeled split $\mathcal{P}^{U}$  (without using the original labels), with each containing 399 sequences. Then, we randomly sample 5\%, 10\%, 20\% and 50\% sequences from $\mathcal{P}^{L}$, which lead to the ratio of labeled data and unlabeled data $\mathcal{P}^{L}:\mathcal{P}^{U}$ as 1:20, 1:10, 1:5 and 1:2, respectively. mAP and mAPH under LEVEL\_2 metric are used to evaluate the 3D object detection performance on the full validation set, where 3D IoU thresholds for ``Vehicle'', ``Pedestrian'' and ``Cyclist'' are 0.7, 0.5 and 0.5, respectively.


\subsection{Implementation Details}\label{subsec:detail}
For the STE module in the teacher model, we have empirically defined the fixed (weak) augmentation types as rotation and double flip. Grid search is conducted to find the more effective rotation parameters, \ie, $\{0\degree, +22.5\degree, -22.5\degree\}$. The final number of augmented views $K$ is computed by the product of flip times and rotation times, \ie, $4\times3=12$ in our case. The strong augmentation includes \textit{random} rotation and flip and scaling, which is the same as that in~\cite{Second_2018Yan}. For the CBV module, we offline train the RoI network~\cite{Deng2021VoxelRT} for 10 epochs based only on the labeled data, with a stop-gradient operation to detach from the backbones. We define positive and negative box samples according to the boxes classification scores and the IoU between ground truths. For example, boxes with classification scores below 0.1 or IoU below 0.3 are treated as negatives, and other boxes are viewed as positives. The votes objectness is optimized by a sigmoid focal loss and the votes localization is optimized by smooth-$\ell_1$ loss. The IoU score for clustering boxes is fixed as 0.5. For the semi-supervised training configuration, we follow the Noisy Student implementation provided by ONCE official benchmark \cite{mao2021one}, \ie, a model pre-trained on the full training set is used to warm up both the student and teacher model. Then the student is trained for 25 to 75 epochs depending on the amount of unlabeled data with learning rate 0.001, and the teacher is updated every 25 epochs. 

\begin{table}[t]
\centering
\caption{\textbf{Evaluation results on ONCE validation set} with different amounts of unlabeled samples (\eg, ``Small'', ``Medium'' and ``Large'') following the official implementation in ONCE \cite{mao2021one}. For better understanding, the best overall result in each class has been highlighted in \textbf{bold} and the relative gains of each SSL method compared to the baseline model (\ie, SECOND \cite{Second_2018Yan} trained with only labeled samples) have been illustrated in colors where the positive gains in \B{\textbf{blue}} and negative gains are in \G{\textbf{green}}. }
\label{tab:comparison_3_splits}
\resizebox{0.98\textwidth}{!}{%
\begin{tabular}{rccccccccccccc}
\hline
\multicolumn{1}{r|}{\multirow{2}{*}{\textbf{Methods}}} &
  \multicolumn{4}{c|}{\textbf{Vehicle AP } (\%)} &
  \multicolumn{4}{c|}{\textbf{Pedestrian AP } (\%)}  &
  \multicolumn{4}{c|}{\textbf{Cyclist AP } (\%)} &
  \multicolumn{1}{c}{\multirow{2}{*}{\textbf{mAP} (\%)}} \\
\multicolumn{1}{r|}{} &  overall &  0-30m &  30-50m &
  \multicolumn{1}{l|}{50m-inf} &  overall &  0-30m &  30-50m &
  \multicolumn{1}{l|}{50m-inf} &  overall &  0-30m &  30-50m &  
  \multicolumn{1}{l|}{50m-inf} &  \multicolumn{1}{c}{} \\ \hline
\multicolumn{1}{r|}{Baseline \cite{Second_2018Yan}} &  71.19 &  84.04 &  63.02 &
  \multicolumn{1}{l|}{47.25} &  26.44 &  29.33 &  24.05 &
  \multicolumn{1}{l|}{18.05} &  58.04 &  69.96 &  52.43 &
  \multicolumn{1}{l|}{34.61} &  51.89 \\ \hline
\multicolumn{14}{c}{\textbf{Small} (100K unlabeled Samples)} \\ \hline
\multicolumn{1}{r|}{Pseudo Label} &  72.80 &  84.46 &  64.97 &
  \multicolumn{1}{l|}{51.46} &  25.50 &  28.36 &  22.66 &
  \multicolumn{1}{l|}{18.51} &  55.37 &  65.95 &  50.34 &
  \multicolumn{1}{l|}{34.42} &  51.22 (\textbf{\G{- 0.67}}) \\
\multicolumn{1}{r|}{Noisy Student \cite{Noisy_student_2020}} &  73.69 &  84.69 &  67.72 &
  \multicolumn{1}{l|}{53.41} &  28.81 &  33.23 &  23.42 &
  \multicolumn{1}{l|}{16.93} &  54.67 &  65.58 &  50.43 &
  \multicolumn{1}{l|}{32.65} &  52.39 (\textbf{\B{+ 0.50}}) \\
\multicolumn{1}{r|}{Mean Teacher \cite{Mean_Teacher_2017_Tarvainen}} &  74.46 &  86.65 &  68.44 &
  \multicolumn{1}{l|}{53.59} &  30.54 &  34.24 &  26.31 &
  \multicolumn{1}{l|}{20.12} &  61.02 &  72.51 &  55.24 &
  \multicolumn{1}{l|}{39.11} &  55.34 (\textbf{\B{+ 3.45}})\\
\multicolumn{1}{r|}{SESS \cite{SESS_2020}} &  73.33 &  84.52 &  66.22 &
  \multicolumn{1}{l|}{52.83} &  27.31 &  31.11 &  23.94 &
  \multicolumn{1}{l|}{19.01} &  59.52 &  71.03 &  53.93 &
  \multicolumn{1}{l|}{36.68} &  53.39 (\textbf{\B{+ 1.50}})\\
\multicolumn{1}{r|}{3DIoUMatch \cite{3DIoUMatch_2021}} &  73.81 &  84.61 &  68.11 &
  \multicolumn{1}{l|}{54.48} &  30.86 &  35.87 &  25.55 &
  \multicolumn{1}{l|}{18.30} &  56.77 &  68.02 &  51.80 &
  \multicolumn{1}{l|}{35.91} &  53.81 (\textbf{\B{+ 1.92}})\\ 
\multicolumn{1}{r|}{\textbf{Our Mehtod}} & \textbf{76.07}  &{86.78}   &{70.19}   &
  \multicolumn{1}{l|}{{56.17}} & \textbf{35.90}  & {39.98}  & {31.67}  & 
  \multicolumn{1}{l|}{{24.37}} & \textbf{61.19}  &{73.97}   & {55.13}  & 
  \multicolumn{1}{l|}{{36.98}} & \textbf{57.72 (\B{+ 5.83})}\\ \hline
  
\multicolumn{14}{c}{\textbf{Medium} (500K unlabeled Samples)} \\ \hline
\multicolumn{1}{r|}{Pseudo Label} &  73.03 &  86.06 &  65.96 &
  \multicolumn{1}{l|}{51.42} &  24.56 &  27.28 &  20.81 &
  \multicolumn{1}{l|}{17.00} &  53.61 &  65.26 &  48.44 &
  \multicolumn{1}{l|}{33.58} &  50.40 (\textbf{\G{- 1.49}})\\
\multicolumn{1}{r|}{Noisy Student \cite{Noisy_student_2020}} &  75.53 &  86.52 &  69.78 &
  \multicolumn{1}{l|}{55.05} &  31.56 &  35.80 &  26.24 &
  \multicolumn{1}{l|}{21.21} &  58.93 &  69.61 &  53.73 &
  \multicolumn{1}{l|}{36.94} &  55.34 (\textbf{\B{+ 3.45}})\\
\multicolumn{1}{r|}{Mean Teacher \cite{Mean_Teacher_2017_Tarvainen}} &  76.01 &  86.47 &  70.34 &
  \multicolumn{1}{l|}{55.92} &  35.58 &  40.86 &  30.44 &
  \multicolumn{1}{l|}{19.82} &  63.21 &  74.89 &  56.77 &
  \multicolumn{1}{l|}{40.29} &  58.27 (\textbf{\B{+ 6.38}})\\
\multicolumn{1}{r|}{SESS \cite{SESS_2020}} &  72.11 &  84.06 &  66.44 &
  \multicolumn{1}{l|}{53.61} &  33.44 &  38.58 &  28.10 &
  \multicolumn{1}{l|}{18.67} &  61.82 &  73.20 &  56.60 &
  \multicolumn{1}{l|}{38.73} &  55.79 (\textbf{\B{+ 3.90}})\\
\multicolumn{1}{r|}{3DIoUMatch \cite{3DIoUMatch_2021}} &  75.69 &  86.46 &  70.22 &
  \multicolumn{1}{l|}{56.06} &  34.14 &  38.84 &  29.19 &
  \multicolumn{1}{l|}{19.62} &  58.93 &  69.08 &  54.16 &
  \multicolumn{1}{l|}{38.87} &  56.25 (\textbf{\B{+ 4.36}})\\
\multicolumn{1}{r|}{\textbf{Our Mehtod}} &  \textbf{78.07} &  {87.43} &\textbf{72.5}   & 
  \multicolumn{1}{l|}{{59.51}} &  \textbf{38.38}  & {42.45}  & {34.62}   & 
  \multicolumn{1}{l|}{{25.58}} &  \textbf{63.23}  & {74.70} & {58.19}  & 
  \multicolumn{1}{l|}{{40.73}} &  \textbf{59.89 (\B{+ 8.00})}\\ \hline
  
\multicolumn{14}{c}{\textbf{Large} (1M unlabeled Samples)} \\ \hline
\multicolumn{1}{r|}{Pseudo Label} &  72.41 &  84.06 &  64.54 &
  \multicolumn{1}{l|}{50.05} &  23.62 &  26.80 &  20.13 &
  \multicolumn{1}{l|}{16.66} &  53.25 &  64.69 &  48.52 &
  \multicolumn{1}{l|}{33.47} &49.76 (\textbf{\G{- 2.13}})\\
\multicolumn{1}{r|}{Noisy Student \cite{Noisy_student_2020}} &  75.99 &  86.67 &  70.48 &
  \multicolumn{1}{l|}{55.60} &  33.31 &  37.81 &  28.19 &
  \multicolumn{1}{l|}{21.39} &  59.81 &  70.01 &  55.13 &
  \multicolumn{1}{l|}{38.33} &  56.37 (\textbf{\B{+ 4.48}})\\
\multicolumn{1}{r|}{Mean Teacher \cite{Mean_Teacher_2017_Tarvainen}} &  76.38 &  86.45 &  70.99 &
  \multicolumn{1}{l|}{57.48} &  35.95 &  41.76 &  29.05 &
  \multicolumn{1}{l|}{18.81} &  \textbf{65.50} &  75.72 &  60.07 &
  \multicolumn{1}{l|}{43.66} &  59.28 (\textbf{\B{+ 7.39}})\\
\multicolumn{1}{r|}{SESS \cite{SESS_2020}} &  75.95 &  86.83 &  70.45 &
  \multicolumn{1}{l|}{55.76} &  34.43 &  40.00 &  27.92 &
  \multicolumn{1}{l|}{19.20} &  63.58 &  74.85 &  58.88 &
  \multicolumn{1}{l|}{39.51} &  57.99 (\textbf{\B{+ 6.10}})\\
\multicolumn{1}{r|}{3DIoUMatch \cite{3DIoUMatch_2021}} &  75.81 &  86.11 &  71.82 &
  \multicolumn{1}{l|}{57.84} &  35.70 &  40.68 &  30.34 &
  \multicolumn{1}{l|}{21.15} &  59.69 &  70.69 &  54.92 &
  \multicolumn{1}{l|}{39.08} &  57.07 (\textbf{\B{+ 5.18}})\\ 
\multicolumn{1}{r|}{\textbf{Our Mehtod}} & \textbf{78.12}  &87.22   &72.74   &
  \multicolumn{1}{l|}{59.58} & \textbf{41.95}  &48.09   &35.13   &
  \multicolumn{1}{l|}{26.01} &  {64.12} &75.85   &58.04   &
  \multicolumn{1}{l|}{41.45} & \textbf{61.40 (\B{+ 9.51 })}  \\ \hline
\end{tabular}%
}
\end{table}

\let\thefootnote\relax\footnotetext{$^{1}$ https://once-for-auto-driving.github.io/benchmark.html\#benchmark}

\subsection{Main Results} \label{subsec:main_res}

\noindent\textbf{ONCE Results.} First of all, we aim to compare our \textit{ProficientTeachers} with other state-of-the-art SSL approaches on the ONCE dataset \cite{mao2021one}. Here, we borrow all the evaluation results from the official benchmark$^{1}$ \cite{mao2021one}, where five typical SSL approaches have been included, \ie, {Pseudo Label}, {Noisy Student} \cite{Noisy_student_2020}, {Mean Teacher} \cite{Mean_Teacher_2017_Tarvainen}, {SESS} \cite{SESS_2020}, {3DIoUMatch} \cite{3DIoUMatch_2021}. For a fair comparison, the {SECOND} \cite{Second_2018Yan} detector trained with only the labeled samples has been adopted as the baseline. All the comparison results are given in \tabref{tab:comparison_3_splits}. Compared to the baseline model that is trained with only the labeled samples, all these SSL frameworks can obtain positive gains, with the help of large amounts of unlabeled samples except for the {Pseudo Label} method. This may be because the official implementation of {Pseudo Label} includes no augmentation when training the student. Interestingly, the improvements increase gradually with the increase of the number of unlabeled samples. Compared to the other SSL methods, our framework achieves the best performance among all the three splits, which obtains {5.83}, {8.00}, {9.51} mAP improvements, respectively. 

Since our method removes the necessity of threshold selection, we compare it with the confidence-based filtering method, \ie, Noisy Student~\cite{Noisy_student_2020} with different pseudo-label threshold $c$. In our implementation, pseudo boxes with scores above $c$ are viewed as positive samples, meanwhile, we also ignore the pseudo boxes with scores below $c$ to avoid taking potential true positives as negatives. As shown in Fig.~\ref{fig:threshold}  (c), $c=0.3$ achieves better performance (55.75 mAP), which is even stronger than Mean Teacher~\cite{Mean_Teacher_2017_Tarvainen}, while our method still surpasses it a lot.

\begin{table}[t]%
\centering
\parbox{0.45\textwidth}{
\begin{footnotesize}

	\caption{\textbf{Generalizability on different detectors} with our SSL method.}
	\label{tab:evaluation_other_Detectors}
\resizebox{0.48\textwidth}{!}
	{%
		\begin{tabular}{r |c|c| c ccc }
			\hline
			\multicolumn{1}{r| }{\multirow{2}{*}{\textbf{Methods}}}&
			\multicolumn{1}{c| }{\multirow{2}{*}{$\textbf{mAP}$($\%$)}}&  
			\multicolumn{1}{c| }{\multirow{2}{*}{$\textbf{Gain}$}} &
			
			\multicolumn{3}{c}{$\textbf{AP}$ ($\%$)} \\
			\multicolumn{1}{c|}{}& 
			\multicolumn{1}{c|}{}& 
			\multicolumn{1}{c|}{}& 
			\multicolumn{1}{l}{Vehicle} & 
			\multicolumn{1}{l}{Pedestrian} & 
			\multicolumn{1}{l}{Cyclist} \\ \hline
			SECOND \cite{Second_2018Yan} &  51.89 & - & 71.19 &	26.44 &	58.04  \\ 
			Our Method &  59.89 & \textbf{+8.00}  &	78.07 &	38.38 &	63.23 \\ \hline
			PV-RCNN \cite{PVRCNN_2020_Shi}  &57.24 & -  & 79.35 & 29.64 &	62.73  \\ 
			Our Method & 63.40 & \textbf{+6.16} &	81.09 &	41.55 &	67.57 \\ \hline
			CenterPoint \cite{CENTERPOINT_2021} & 62.99 &- & 75.26 &	51.65 &	65.79  \\ 
			Our Method & 68.22 &\textbf{+5.23} &	77.77 &	56.34 &	70.55\\ \hline

		\end{tabular}
	}

\end{footnotesize}
}
\qquad
\parbox{0.45\textwidth}{
\begin{footnotesize}

\caption{\textbf{Performance on the the ONCE test set.}}
 	\label{tb:test}
\resizebox{0.45\textwidth}{!}
 	{
 		\begin{tabular}{r |c| ccc }
 			\hline
 			\multicolumn{1}{r| }{\multirow{2}{*}{$\textbf{Methods}$}}&
 			\multicolumn{1}{c| }{\multirow{2}{*}{$\textbf{mAP}$ ($\%$)}}&
 			\multicolumn{3}{c}{$\textbf{AP}$ ($\%$)} \\
 			\multicolumn{1}{c|}{}& 
 			\multicolumn{1}{c|}{}& 
 			\multicolumn{1}{l}{Vehicle} & 
 			\multicolumn{1}{l}{Pedestrian} & 
 			\multicolumn{1}{l}{Cyclist} \\ \hline
 			SECOND \cite{Second_2018Yan}& 51.90  & 69.71 &	26.09 &	59.92  \\  \hline

 			Pseudo Label & 49.29 & 70.29 &	21.85 &	55.72  \\ 
 			Noisy Student~\cite{Noisy_student_2020} & 56.61 & 74.50 &	33.28 &	62.05 \\ 
 			Mean Teacher~\cite{Mean_Teacher_2017_Tarvainen} & 59.99  &	76.60 &	36.37 &	66.99 \\ 
 			SESS~\cite{SESS_2020} & 58.78  &	74.52 &	36.29 &	65.52 \\ 
 			3DIoUMatch~\cite{3DIoUMatch_2021} & 57.43 & 74.48 & 35.74 & 62.06 \\ \hline
 			{Our Method} & { 61.44 } &{ 76.85 } &	{ 41.27 } &	{ 66.19 } \\
 			 			
 			\hline
 		\end{tabular}
 	}
\end{footnotesize}

}
\end{table}

We also verify our SSL method on two more baselines, which are PV-RCNN \cite{PVRCNN_2020_Shi} and CenterPoint \cite{CENTERPOINT_2021}. PV-RCNN is a hybrid point-voxel approach that inherits the advantages from both point \cite{PointNet++_2017} and voxels features \cite{VoxelNet_2018_Zhou}. CenterPoint is a voxel-based anchor-free framework that has achieved SOTA results in several benchmarks. All the methods are trained with the ``Medium'' unlabeled subset. From \tabref{tab:evaluation_other_Detectors}, we can see that our method works well on these different detectors, \eg, still yielding 5.23 points improvements over the strong detector CenterPoint. This indicates the good generalizability of our SSL method.

The results on the testing split of the ONCE benchmark are presented in Table.~\ref{tb:test}, where the results of other SSL methods are reported in terms of the official benchmark. All the SSL methods are based on the SECOND detector for a fair comparison. Among all the competitors, our method shows remarkable superiority. It outperforms the strong competitor Mean Teacher by 1.45 points, and also exceeds the recently proposed SESS and 3DIoUMatch by a large margin.

\noindent\textbf{Waymo Results.} For evaluating on Waymo, we compare our model with the strongest competitor verified on ONCE, \ie, Noisy Student~\cite{Noisy_student_2020} with a carefully selected threshold $c=0.3$. Since this filtering strategy derives from Fixmatch~\cite{FixMatch_2020fixmatch}, we name it Fixmatch for simplicity. The {SECOND} \cite{Second_2018Yan} detector trained with different amounts of labeled data is used as the baseline. As shown in Table~\ref{tb:waymo}, impressive results are obtained. Both Fixmatch and our \textit{ProficientTeachers} achieve better results than the full-supervised baseline, which proves the advantage of semi-supervised learning. In particular, Fixmatch surpasses the baseline by 1.90 to 2.95 mAPH. By contrast, our method obtains much better results than Fixmatch, \ie, improving the baseline by 4.34 to 5.35 mAPH. Moreover, we also run an oracle model that trains the detector with the full 158,361 labels, while our model with only half labels outperforms it. This further demonstrates the generalization of our method over different datasets.

\begin{table}[bt]
\caption{\small{ \textbf{Semi-supervised 3D Object Detection on Waymo dataset.} We train the SECOND~\cite{Second_2018Yan} baseline with different fractions of labeled data. Then, we compare our \textit{ProficientTeachers} model with a strong competitor, Fixmatch~\cite{FixMatch_2020fixmatch}. It shows that our model can consistently improve the detection performance.}}
\label{tb:waymo}
\centering
\renewcommand\arraystretch{1.00}
\resizebox{0.98\textwidth}{!}{
\begin{tabular}{l||c||c|cccc}
\hline\thickhline
\rowcolor[HTML]{EFEFEF} ~~~~~~~\textbf{Different } 
\cellcolor[HTML]{EFEFEF}                     & &\textbf{Performance} &\multicolumn{4}{c}{\cellcolor[HTML]{EFEFEF}~~~~\textbf{3D AP/APH @0.7 (LEVEL 2)}}  \\
\rowcolor[HTML]{EFEFEF} 
\multirow{-2}{*}{\cellcolor[HTML]{EFEFEF}} ~ \textbf{Label Amounts} &\multirow{-2}{*}{\textbf{Training Paradigm} }
&\textbf{ Gain} & \textbf{Overall}             & \textbf{Vehicle}           &\textbf{ Pedestrian}            & \textbf{Cyclist}         \\ \hline

     & Baseline~\cite{Second_2018Yan}  & -/-  & 45.78/40.40          & 50.03/49.52           & 45.77/34.98           &41.53/36.69       \\
\multirow{-2}{*}{~~~5\% ($\sim$ 4k Labels)}  &{Fixmatch~\cite{FixMatch_2020fixmatch}} &{{{+3.02/+2.95}} } & 48.80/43.35   & 51.87/51.27   & 48.28/36.56   & 46.26/42.21   \\  \multirow{-1.5}{*}{~~~$\mathcal{P}^{L}:\mathcal{P}^{U}=1:20$}   & {~~ProficientTeacher}&{{\textbf{+5.32/+5.35}} } & 51.10/45.75    & 53.04/52.54   & 50.33/38.67   & 49.92/46.03   \\ \hline

     & Baseline~\cite{Second_2018Yan}  & -/-  & 50.00/45.83           & 54.90/54.25   & 48.45/38.44   & 46.66/44.79 \\
\multirow{-2}{*}{~~~10\% ($\sim$ 8k Labels)}  &{Fixmatch~\cite{FixMatch_2020fixmatch}} &{{{+2.28/+1.90}} } & 52.28/47.73   & 56.60/55.99   & 51.60/40.63   & 48.63/46.56   \\  \multirow{-1.5}{*}{~~~$\mathcal{P}^{L}:\mathcal{P}^{U}=1:10$}   & {~~ProficientTeacher}&{{\textbf{+5.01/+4.60}} } & 55.01/50.43    & 57.59/56.92   & 54.28/43.19   & 53.15/51.18   \\ \hline

     & Baseline~\cite{Second_2018Yan}  & -/-  & 53.09/49.11           & 57.40/56.81           & 51.54/41.91           &50.33/48.62       \\
\multirow{-2}{*}{~~~20\% ($\sim$ 16k Labels)}  &{Fixmatch~\cite{FixMatch_2020fixmatch}} &{{{+2.72/+2.34}} } & 55.81/51.45   & 58.94/58.37   & 54.37/44.23   & 54.11/51.75   \\  \multirow{-1.5}{*}{~~~$\mathcal{P}^{L}:\mathcal{P}^{U}=1:5$}   & {~~ProficientTeacher}&{{\textbf{+5.50/+5.05}} } & 58.59/54.16    & 59.97/59.36   & 57.88/46.97   & 57.93/56.15   \\ \hline

     & Baseline~\cite{Second_2018Yan}  & -/-  & 57.07/53.26           & 60.93/60.37           & 55.98/46.68           &54.31/52.74       \\
\multirow{-2}{*}{~~~50\% ($\sim$ 40k Labels)}  &{Fixmatch~\cite{FixMatch_2020fixmatch}} &{{{+2.72/+2.45}} } & 59.79/55.71 & 61.88/61.34   & 58.64/49.00   & 58.85/56.78      \\  \multirow{-1.5}{*}{~~~$\mathcal{P}^{L}:\mathcal{P}^{U}=1:2$}   & {~~ProficientTeacher}&{{\textbf{+4.57/+4.34}} } & 61.64/57.60    & 63.06/62.50   & 61.53/51.33   & 60.33/58.97   \\ \hline

     & Baseline~\cite{Second_2018Yan}  & -/-  & 59.63/55.94           & 62.78/62.24           & 59.45/50.44           &56.67/55.13       \\
\multirow{1}{*}{~~~100\% ($\sim$ 80k Labels)}  &{Fixmatch~\cite{FixMatch_2020fixmatch}} &{{{+2.43/+2.02}} } & 62.06/57.96 & 63.50/62.98   & 62.00/52.52   & 60.69/58.37      \\  \multirow{2}{*}{~~~$\mathcal{P}^{L}:\mathcal{P}^{U}=1:1$}   & {~~ProficientTeacher}&{{\textbf{+3.33/+3.20}} } & 62.96/59.14    & 63.56/63.06   & 62.34/53.19   & 62.97/61.18   \\ \cline{2-7}
 & {~~Oracle Model~\cite{Second_2018Yan} }&{{{+2.80/+2.93}} } & 62.43/58.87     & 65.43/64.91   & 61.82/52.93  & 60.03/58.76   \\

 \hline

 \thickhline
  
\end{tabular}}
\end{table}

\begin{figure}[t]
\includegraphics[width=0.98\textwidth]{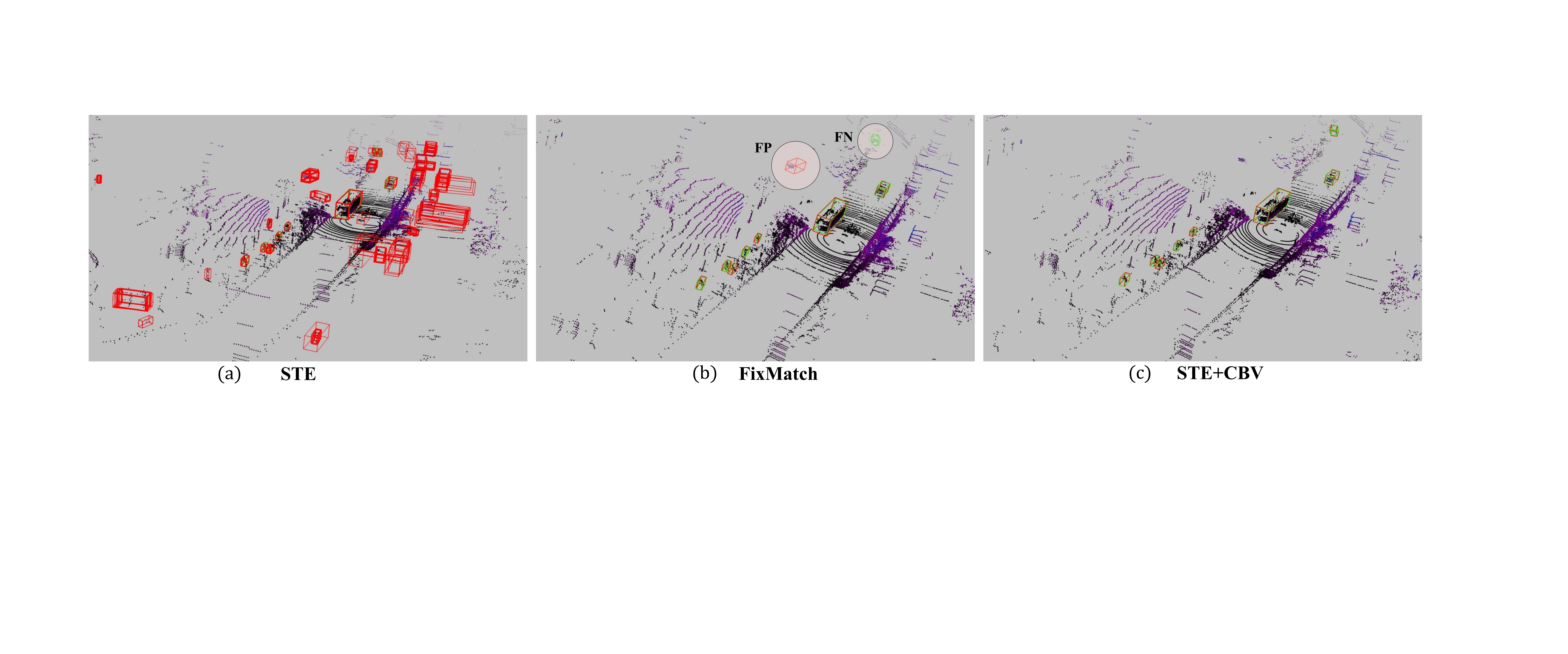}
\caption{\textbf{Visualization of pseudo labels} produced by our STE and CBV modules, or by threshold-based FixMatch~\cite{FixMatch_2020fixmatch}. Predictions are in red and GTs are in green.}

\label{fig:visual}
\end{figure}

\noindent\textbf{Qualitative Results.} We visualize some examples of pseudo labels on ONCE in Fig. \ref{fig:visual}. The false positives and false negatives are highlighted in circles. We compare our STE and CBV modules with the threshold-based method FixMatch~\cite{FixMatch_2020fixmatch}. It incurs FN when detecting distant objects with sparse points, and it gives FP when detecting a hard distractor as shown in the circles in (b) (plz zoom in for a better view). In contrast, by gathering boxes from different views, our STE successfully addresses the FN object. As for the FP box, its score will be refined and re-scaled in our CBV such that a box with low detection frequency will be removed accordingly. Moreover, for boxes that are close to GTs, their box coordinates will also be refined after voting in a cluster. All these designs lead to high-quality pseudo labels.

\subsection{Ablation Studies}\label{subsec:ablation}
A series of ablation studies are set to verify the effectiveness of different modules and all the results are given in \tabref{tab:ablation_study}. All the experiments are conducted based on the SECOND detector with the ``Small'' unlabeled subset. Our \textit{ProficientTeachers} model is mainly based on the Noisy Student implementation in ONCE, and the default confidence threshold $c$ for filtering the noisy pseudo labels is 0.1. To explore a more effective threshold, we perform grid search and find that $c=0.3$ gives better detection results, as shown in Fig.~\ref{fig:threshold} (c). Thus we use this threshold for the subsequent experiments. STE, CBV and BCL are the three necessary modules proposed in this work, and we ablate the contribution of each module. First, the STE module has been used to generate more pseudo boxes. We evaluate it by using $c=0.3$ to filter the boxes. This improves the baseline by 0.46 points. Next, we replace the confidence-based thresholding with the proposed CBV module to adaptively aggregate these pseudo boxes. This exceeds the baseline by 1.42 points. Finally, thanks to the BCL module, our full model achieves 57.72\% mAP, improving the baseline by 1.97 points.


\begin{table}[t]
\centering
	\caption{\textbf{Ablation studies to verify the effect of different modules.} The experiments are conducted based on the SECOND with the ``Small'' unlabeled subset.}
	\label{tab:ablation_study}
	\resizebox{0.95\textwidth}{!}
	{%
		\begin{tabular}{c |c| c|c|ccc }
			\hline
			\multicolumn{1}{c| }{\multirow{2}{*}{\textbf{Module}}}& 
			\multicolumn{1}{c| }{\multirow{2}{*}{\textbf{Aspect}}}&
			\multicolumn{1}{c| }{\multirow{2}{*}{$\textbf{mAP}$ ($\%$)}}&
			\multicolumn{1}{c| }{\multirow{2}{*}{$\textbf{Gain}$}} &
			\multicolumn{3}{c}{$\textbf{AP}$ ($\%$)} \\
			\multicolumn{1}{c|}{}& 
			\multicolumn{1}{c|}{}&
			\multicolumn{1}{c|}{}& 
			\multicolumn{1}{c|}{}& 
			\multicolumn{1}{l}{Vehicle} & 
			\multicolumn{1}{l}{Pedestrian} & 
			\multicolumn{1}{l}{Cyclist} \\ \hline
		   {Noisy Student (Baseline) } & c = 0.3& 55.75 &  -   &74.66 &	32.25 &	60.34 \\  \hline
		   
		     BCL Module & c = 0.3, \textit{w/o} STE or CBV  & 56.54 &\textbf{+0.79}  & 75.70	 &	33.65 &	60.28  \\ \hline
			STE Module  &c = 0.3, \textit{w/o}  BCL or CBV & 56.21 & \textbf{+0.46}  & 75.62 &  33.14 &	59.87  \\ 
			 STE+CBV Modules & \textit{w/o} BCL   & 57.17 &\textbf{+1.42}  & 	75.98 &	34.86 &	60.67 \\   \hline

			Full Model & \textit{w/} STE, CBV and BCL & 57.72  &\textbf{+1.97}  &	76.07 &	35.90 &	61.19\\ \hline

		\end{tabular}
	}
\end{table}

\section{Discussion and Conclusion}
In this work, we proposed a new SSL framework for LiDAR-based 3D object detection. In particular, our work focuses on improving the FP and FN in the pseudo labels produced by the teacher model. First, to address the FN, a spatial-temporal ensemble (STE) module is introduced to produce sufficient seed boxes and ensure a high recall. This is realized by a spatial data augmentation and a temporal model ensemble. Second, to resolve the FP predictions and improve the precision, we developed a clustering-based box voting (CBV) module that performs voting and aggregating based on boxes in a cluster. More importantly, our CBV can yield high-quality pseudo labels without the need of deliberately selecting thresholds. The STE and CBV modules enhance the original teacher to {proficient teachers}. Finally, we proposed a box-wise contrastive learning (BCL) strategy to optimize the student towards cross-view feature consistency, reducing the effect of inaccurate pseudo labels. Experiments on the large-scale ONCE and Waymo datasets demonstrated the superiority of our method.

\noindent\textbf{Acknowledgements.}
This work was partially supported by Zhejiang Lab’s International Talent Fund for Young Professionals (ZJ2020GZ023), ARC DECRA DE220101390, FDCT under grant 0015/2019/AKP, and the Start-up Research Grant (SRG) of University of Macau.

\clearpage
%
%
\bibliographystyle{splncs04}
\bibliography{3655.bib}
\end{document}